\documentclass[10pt,twocolumn,letterpaper]{article}

\pdfoutput=1

\usepackage[
]{cvpr}      % To produce the REVIEW version
\usepackage{graphicx}
\usepackage{amsmath}
\usepackage{amssymb}
\usepackage{booktabs}

\usepackage[noend]{algpseudocode}
\usepackage{algorithmicx,algorithm}

\usepackage{balance} % for balancing columns on the final page

\usepackage[pagebackref,breaklinks,colorlinks]{hyperref}

\usepackage[capitalize]{cleveref}
\crefname{section}{Sec.}{Secs.}
\Crefname{section}{Section}{Sections}
\Crefname{table}{Table}{Tables}
\crefname{table}{Tab.}{Tabs.}

\def\cvprPaperID{11269} % *** Enter the CVPR Paper ID here
\def\confName{CVPR}
\def\confYear{2023}

\begin{document}

\title{Actor-Director-Critic: A Novel Deep Reinforcement Learning Framework}

\author{Zongwei Liu\\
Xi’an Jiaotong University\\
China\\
{\tt\small Ting\_Haode@stu.xjtu.edu.cn}
\and
Yonghong Song\\
Xi’an Jiaotong University\\
China\\
{\tt\small songyh@xjtu.edu.cn}
\and
Yuanlin Zhang\\
Xi’an Jiaotong University\\
China\\
{\tt\small ylzhangxian@xjtu.edu.cn}
}
\maketitle

%%%%%%%%% ABSTRACT
\begin{abstract}
   In this paper, we propose actor-director-critic, a new framework for deep reinforcement learning. Compared with the actor-critic framework, the director role is added, and action classification and action evaluation are applied simultaneously to improve the decision-making performance of the agent. Firstly, the actions of the agent are divided into high quality actions and low quality actions according to the rewards returned from the environment. Then, the director network is trained to have the ability to discriminate high and low quality actions and guide the actor network to reduce the repetitive exploration of low quality actions in the early stage of training. In addition, we propose an improved double estimator method to better solve the problem of overestimation in the field of reinforcement learning. For the two critic networks used, we design two target critic networks for each critic network instead of one. In this way, the target value of each critic network can be calculated by taking the average of the outputs of the two target critic networks, which is more stable and accurate than using only one target critic network to obtain the target value. In order to verify the performance of the actor-director-critic framework and the improved double estimator method, we applied them to the TD3 algorithm to improve the TD3 algorithm. Then, we carried out experiments in multiple environments in MuJoCo and compared the experimental data before and after the algorithm improvement. The final experimental results show that the improved algorithm can achieve faster convergence speed and higher total return.
\end{abstract}
\section{Introduction}
\label{sec:intro}
Deep reinforcement learning is a trial-and-error approach to learning, which takes the data obtained from the interaction between the agent and the environment as samples for the agent's policy learning. In the model-free reinforcement learning algorithm, the agent's cognition of the environment comes entirely from the sample data obtained during the interaction. It is this iterative trial-and-error learning method that makes a high percentage of low-quality data in a large amount of sample data, resulting in a poor utilization of the samples. Therefore, if there is a way to improve the sample utilization, the learning speed of reinforcement learning can be improved. There are many excellent existing model-free reinforcement learning algorithms that combine policy functions and value functions based on the actor-critic framework, which are applicable to a variety of action type problems with improved sample training efficiency. In the actor-critic framework, actor learns parameterized policy function and critic learns parameterized value function. The learned value function can provide more feedback information to the policy function than the environmental reward to guide the policy function in action decision making. Therefore, the learning of the policy function depends on the quality of the estimates of the value function that undergoes learning simultaneously. Because of this, it leads to the problem that until the value function can provide a reasonable signal to the policy function, it is difficult for the agent to rely on the policy function to make good actions.

\begin{figure*}[htbp]
	\centering
	\includegraphics[width=0.85\linewidth]{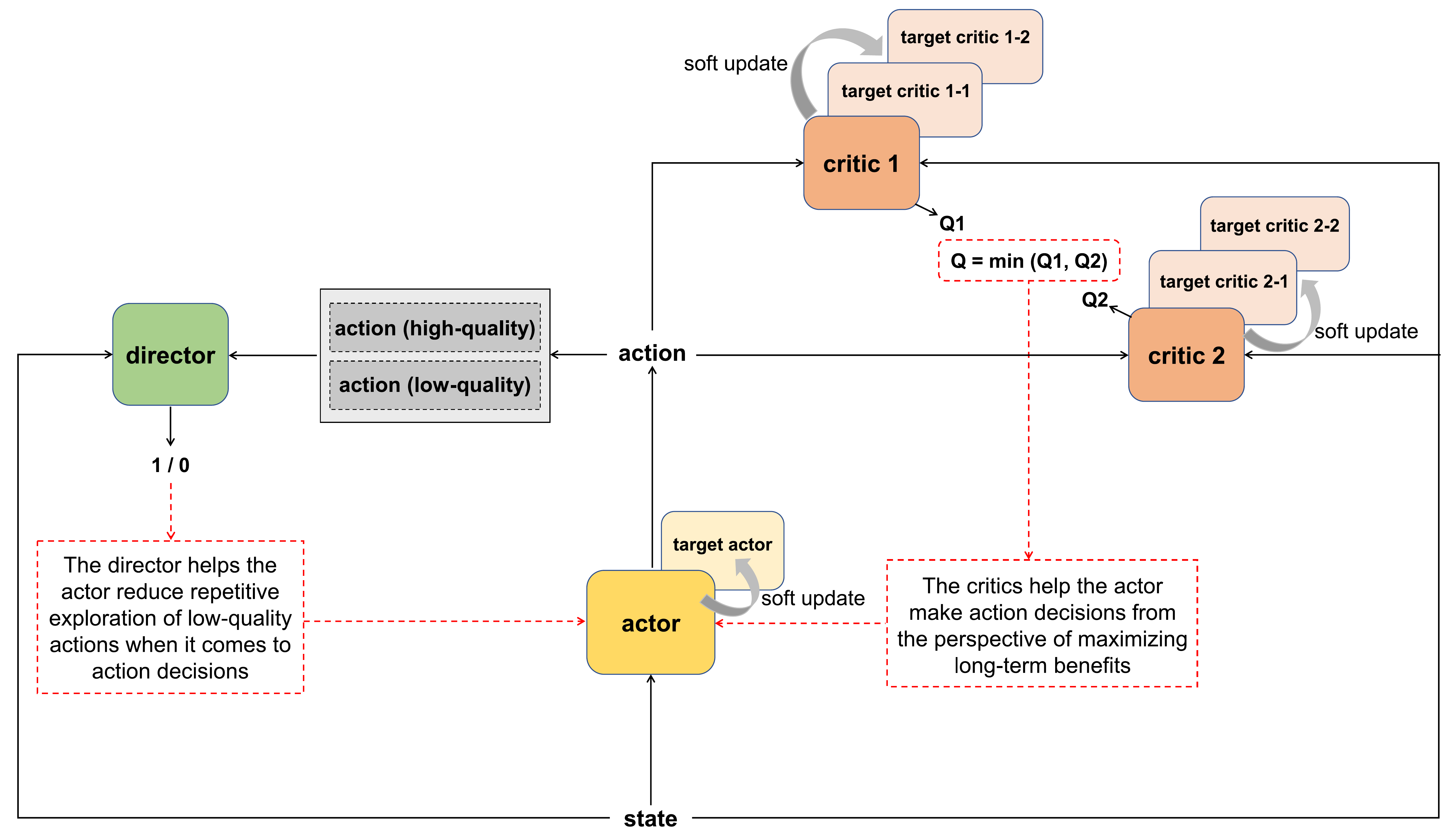}
	\caption{Implementation process of CTD3.}
	\label{fig:short}
\end{figure*}

In order to provide more guidance to the policy function before the value function can provide reasonable guidance signals for the policy function and speed up the learning of the policy function, we propose the actor-director-critic framework. Compared with actor-critic framework, actor-director-critic framework adds a director network with discriminative function to discriminate between low quality and high quality actions. Director can provide guidance other than critic to actor at the early stage of actor training, so that agent can reduce the repetitive exploration of low quality actions in the initial trial-and-error learning process, and further improve the training efficiency of samples. In the process of interacting with the environment, the agent receives a reward back from the environment for each action it performs. We simply classify the actions into two categories based on the reward returned, one is low quality actions with relatively low reward value, and the other is high quality actions with relatively high reward value. Before training the director network, we divided the empirical dataset into two categories, a low quality dataset consisting of low quality actions and their corresponding rewards and states, and a high quality dataset consisting of high quality actions and their corresponding rewards and states. We then use these two datasets above to train the ability of the director network to discriminate between high and low quality actions, so that the director network can help the actor network to reduce the repetitive exploration of low quality actions during policy optimization. We set attenuation factor so that the director focuses on the early stages of actor training.

In addition, there is the problem of overestimation in reinforcement learning. The overestimation problem is usually caused by two reasons. One is the idea of maximization of the objective policy, where the maximum value of the estimated value function is used as the estimate of the true value, which leads to maximization bias and thus overestimation. The second is the bootstrap phenomenon, which means using the predicted value of the value function at the future moment to update the estimated value of the value function at the present moment, which will also bring about the overestimation problem. The double estimator method is a common method to solve the overestimation problem. And existing method learns two independent Q networks, each of which has a target Q network. The target Q network copies its parameters from the Q network at regular intervals, and the output of the target Q network is then considered as the target value to optimize the Q network. The problem is that the output of a single target Q is often unstable, and this instability will reduce the accuracy of the target value and lower the training efficiency.

In order to better solve the overestimation problem, we propose an improved double estimator method, which is to design two target Q networks for each Q network separately, and these two target Q networks update the parameters alternately according to a fixed time interval. The average of the outputs of the two target Q networks is used as the target value when optimizing the Q network. This method will obtain more stable and accurate target values than using a single target Q network, and thus can better solve the overestimation problem.

The TD3 \cite{fujimoto2018addressing} algorithm is an excellent performing algorithm based on the actor-critic framework. To test the performance of the above actor-director-critic framework and the improved double estimator method, we improve the TD3 algorithm using the actor-director-critic framework and the improved double estimator method, and refer to the improved algorithm as CTD3 algorithm for short. We conducted comparison experiments and ablation experiments in several environments in the MuJoCo, and all of them achieved good results.
%------------------------------------------------------------------------
\section{Related work}
\label{sec:formatting}
%-------------------------------------------------------------------------
\subsection{Actor-Critic framework}
Model-free reinforcement learning algorithms can be classified as policy-based algorithms, value-based algorithms, and algorithms that combine policy and value. The agent forms a trajectory $\tau$ during its interaction with the environment, and the policy-based algorithm is to learn an action-generating function called policy function $\pi$, such that the agent is able to maximize the cumulative reward when making decisions under the guidance of this policy function. The cumulative reward $\mathcal J(\tau)$ can be expressed as follows:  
\begin{eqnarray}
	\mathcal J(\tau)=\sum\limits_{k=0}^{K} \gamma^{k} r_{t+k+1}
\end{eqnarray}

The discount factor $\gamma$ represents the effect of future rewards on the present and takes a value between 0 and 1. At different moments, the agent is in different states, so the policy function $\pi$ is meant to produce an action $a=\pi(s)$ with state s as input, indicating the probability of taking action a in the case of state s. The policy-based algorithm directly optimizes the cumulative reward $\mathcal J(\tau)$, which is of most interest to the agent. And policy-based algorithm can be applied to problems of either discrete or continuous action types, with the disadvantage that the sample training is relatively inefficient.
The value function can provide information about $\mathcal J(\tau)$. The value function has two forms, one is the state value function, which indicates the expected value of the subsequent cumulative reward that can be obtained in a certain state s. The state value function $\mathcal V(s)$ can be expressed as follows:
\begin{eqnarray}
	\mathcal V(s)=\mathbb{E}(\sum\limits_{k=0}^{K}\gamma_{k} r_{t+k+1}|S_{t}=s)
\end{eqnarray}

The other is the action value function, which represents the expected value of the cumulative reward that can be obtained after performing action a in some state s. The action value function $Q(s,a)$ can be expressed as follows: 
\begin{eqnarray}
	Q(s,a)=\mathbb{E}(\sum\limits_{k=0}^{K}\gamma_{k} r_{t+k+1}|S_{t}=s,A_{t}=a)
\end{eqnarray}

The value-based algorithm firstly learns a value function, and either the state value function or the action value function ultimately uses the computed (s,a) to generate the optimal policy. The contribution of each action is measured by calculating and comparing the output of the function corresponding to all possible actions in the current state. The agent should be in a state with high value as far as possible and choose actions with high value as far as possible, so as to achieve the purpose of maximizing cumulative rewards. The advantage of the value-based algorithm is that the sample training is more efficient, and the disadvantage is that it is often limited to problems with discrete action types and is not suitable for problems where the target policy is a random one.

The algorithm based on the actor-critic framework is an algorithm that combines the policy gradient method and the value function method, which can not only solve the problem of multiple action types, but also the sample training efficiency is relatively high. Two parameterized functions need to be learned in the actor-critic framework, in which actor learns the parameterized policy function and critic learns the parameterized value function. The policy function guides the agent to make action decisions, and the value function is responsible for evaluating the actions. Actor is a network representing the policy function, which selects the actions to be performed based on the current state information of the environment, and learns a better policy using the policy gradient under the guidance of critic. Actor's optimization objective can be expressed as follows:
\begin{eqnarray}
	max  \mathcal J = \mathbb{E} [Q(s,a)]
\end{eqnarray}

Critic is a network representing the value function that learns a value function based on the data collected during the actor's interaction with the environment, and helps the actor to make policy updates by evaluating the quality of the actor's actions. The optimization objective of critic can be expressed as follows:
\begin{eqnarray}
	min  \mathcal L = \mathbb{E} [(Q(s,a) - (r + \gamma Q(s^{\prime},a^{\prime})))^{2}]
\end{eqnarray}

The learned value function can provide more feedback information to the policy function than the environmental reward, so the policy function can achieve better results when it learns based on the information provided by the value function that has completed learning. In the field of single-agent reinforcement learning, many existing excellent algorithms use the actor-critic framework, such as DDPG \cite{lillicrap2015continuous}, TRPO \cite{schulman2015trust}, A3C \cite{mnih2016asynchronous}, PPO \cite{schulman2017proximal}, ACER \cite{wang2016sample}, SAC \cite{haarnoja2018soft}, D4PG \cite{barth2018distributed}, TD3 \cite{fujimoto2018addressing}, DAC \cite{zhang2019dac}, TAAC \cite{yu2021taac}, DCAC \cite{bouteiller2020reinforcement}, ESAC \cite{suri2022off}, and so on. In the field of multi-agent reinforcement learning, there are also many multi-agent algorithms that continue to use the actor-critic framework, such as MADDPG \cite{lowe2017multi}, COMA \cite{foerster2018counterfactual}, IPPO \cite{de2020independent}, SEAC \cite{christianos2020shared}, CM3 \cite{yang2020cm3}, MACAAC \cite{parnika2021attention}, MAPPO \cite{yu2021surprising}, HAPPO \cite{kuba2021trust}, and so on. In addition, the QMIX \cite{rashid2018qmix} algorithm also borrows ideas from the actor-critic framework, where the agent RNN network is equivalent to the actor network and the mixing network is equivalent to the critic network. Of course, there are also many algorithms that improve on the actor-critic framework, such as the multi-agent algorithm MAAC \cite{iqbal2019actor}, which introduces the attention mechanism based on the actor-critic framework and proposes the actor-attention-critic framework, in the hope that the agent can selectively focus on the information that is more conducive to gaining greater returns when learning other agent policies.

As stated in the introduction, reinforcement learning suffers from relatively low sample utilization and, for the actor-critic framework, it is difficult for an agent to rely on the policy function to make good actions until the value function can provide a reasonable signal for the policy function. Regarding the two roles in the actor-critic framework, we can regard the actor as the student and the critic as the teacher. In real life, a student often needs more teachers to teach him to learn more, and the more teachers a student has, the higher the probability of getting good grades. So the concept of multiple teachers can be introduced on the basis of the actor-critic framework. In the actor-director-critic framework proposed in this paper, the director is equivalent to another teacher for the actor, providing guidance to the actor other than the critic and helping the actor learn better policies faster.

%-------------------------------------------------------------------------
\subsection{Double estimator method}
Double estimator is a common method to solve the overestimation problem, and two methods of double estimator will be described below. One is a method that has been used in some algorithms such as Double Q-Learning \cite{hasselt2010double}, which learns two independent Q networks, and each Q network uses the output value of the other during training as an estimate of the action value function of the optimal action sought by itself. This can be described as follows:
\begin{eqnarray}
	y_{1}=r+\gamma Q_{2}(s^{\prime},a^{\prime})
\end{eqnarray}
\begin{eqnarray}
	y_{2}=r+\gamma Q_{1}(s^{\prime},a^{\prime})
\end{eqnarray}

Although two Q networks are used, only one Q network is updated at each update, and they both have a $50\%$ probability of being selected randomly. Another method, which has been used in some algorithms such as TD3 \cite{fujimoto2018addressing}, also uses two Q networks that are independent of each other. The difference is that both networks are updated simultaneously during the training process, using the minimum of their outputs as the estimate of the value function. Moreover, when updating the network parameters, the estimate of the value function is shared between the two Q networks in the calculation of the loss function. This can be described as follows:
\begin{eqnarray}
	y=r+\gamma min Q_{1,2}(s^{\prime},a^{\prime})
\end{eqnarray}

In the existing actor-critic framework, two relatively independent critic networks need to be trained, and then the risk of overestimation is diluted by randomly selecting the output value of one of the critic networks or by taking the minimum value of the output of the two critic networks. In the two double estimator methods mentioned above, each critic network has only one target critic network. In order to better solve the overestimation problem, this paper proposes an improved double estimator method by designing two target critic networks for each critic network.

%%%%%%%%%%%%%%%%%%%%%%%%%%%%%%%%%%%%%%%%%%%%%%%%%%%%%%%%%%%%%%%%%%%%%%%%
\mathchardef\mhyphen="2D	
\begin{algorithm}
	\caption{CTD3} 	
	\begin{flushleft}
		\hspace*{0.1in}
		Initialize the director network $D_{\phi}$, critical network $Q_{\omega_{1}}$, \\
		\hspace*{0.1in}
		$Q_{\omega_{2}}$ and actor network $\mu_{\theta}$ with random parameters $\phi$, \\
		\hspace*{0.1in}
		$\omega_{1}$, $\omega_{2}$, $\theta$ \\
		\hspace*{0.1in}
		Initialize target networks $\omega_{1 \mhyphen 1}^{\prime} \leftarrow \omega_{1}$, $\omega_{1 \mhyphen 2}^{\prime} \leftarrow \omega_{1}$, \\
		\hspace*{0.1in}
		$\omega_{2 \mhyphen 1}^{\prime} \leftarrow \omega_{2}$, $\omega_{2 \mhyphen 2}^{\prime} \leftarrow \omega_{2}$, $\theta^{\prime} \leftarrow \theta$ \\
		\hspace*{0.1in}
		Initialize three replay buffers $B$, $B_{1}$, $B_{2}$ 
	\end{flushleft}
	\begin{algorithmic}[0]
		\For{t = 1 to T} 	
		\State Select action with exploration noise
		\State \hspace*{0.1in} $\widetilde{a_{t}} = \mu_{\theta}(s_{t})+\epsilon_{t}$, $\epsilon_{t} \leftarrow \mathcal N_{t}(0,\sigma)$ 
		
		\State Execute action $\widetilde{a_{t}}$ to get reward $r_{t}$ and new state $s_{t+1}$
		
		\State Store transition tuple ($s_{t}$, $a_{t}$, $r_{t}$, $s_{t+1}$) in $B$
		
		\State Choose a reward value cut-off of the right size, R
		
		\If{$r_{t}$ $\mathrel >$ R} % If 语句，需要和EndIf对应
		\State $a_{t} \mathrel \in a_{h}$, Store ($s_{t}$, $a_{t}$, $r_{t}$, $s_{t+1}$) in $B_{1}$
		\Else
		\State $a_{t} \mathrel \in a_{l}$, Store ($s_{t}$, $a_{t}$, $r_{t}$, $s_{t+1}$) in $B_{2}$
		\EndIf
		
		\State Sample a random minibatch of N transitions \\
		\hspace*{0.3in}
		($s_{j}$, $a_{j}$, $r_{j}$, $s_{j+1}$) from $B_{1}$
		
		\State Sample a random minibatch of N transitions \\
		\hspace*{0.3in}
		($s_{k}$, $a_{k}$, $r_{k}$, $s_{k+1}$) from $B_{2}$
		
		\State Use ($s_{j}$, $a_{j}$, $r_{j}$, $s_{j+1}$), ($s_{k}$, $a_{k}$, $r_{k}$, $s_{k+1}$) to update\\
		\hspace*{0.17in}
		director network $D_{\phi}$: 
		
		\State \hspace*{0.1 in} $\nabla_{\phi} \mathcal V(\phi) =$ \\
		\hspace*{0.4 in}
		$\frac{1}{N} \sum\limits_{i=1}^{N} (\nabla_{\phi} D_{\phi} (s_{i},a_{hi}) - \nabla_{\phi} D_{\phi} (s_{i},a_{li}))$
		
		\State Sample a random minibatch of N transitions \\
		\hspace*{0.3 in}
		($s_{i}$, $a_{i}$, $r_{i}$, $s_{i+1}$) from $B$
		
		\State \hspace*{0.1in} $\widetilde{a_{i}} = \mu_{\theta}(s_{i})+\epsilon_{i}$,  $\epsilon_{i} \leftarrow \mathcal N_{t}(0,\sigma)$ 
		
		\State \hspace*{0.1in} $\widetilde{a_{i+1}} = \mu_{\theta^{\prime}}(s_{i+1})+\epsilon_{i}^{\prime}$,  
		$\epsilon_{i}^{\prime} \leftarrow clip(\mathcal N_{t}(0,\sigma^{\prime}),\mhyphen c,c)$
		
		\State \hspace*{0.1in} $ y_{i} = r_{i} + $ $ \gamma_{Q} min (\frac{Q_{\omega_{1 \mhyphen 1}^{^{\prime}}}(s_{i+1},\widetilde{a_{i+1}}) + Q_{\omega_{1 \mhyphen 2}^{\prime}}(s_{i+1},\widetilde{a_{i+1}})}{2}, $ \\
		\State \hspace*{1.1 in}
		$ \frac{Q_{\omega_{2 \mhyphen 1}^{\prime}}(s_{i+1},\widetilde{a_{i+1}}) + Q_{\omega_{2 \mhyphen 2}^{\prime}}(s_{i+1},\widetilde{a_{i+1}})}{2}) $
		
		\State Use ($s_{i}$, $a_{i}$, $r_{i}$, $s_{i+1}$) to update critic network \\
		\hspace*{0.17in}
		$Q_{\omega_{1}}$ or $Q_{\omega_{2}}$: 
		\State \hspace*{0.1in} $\nabla_{\omega} \mathcal L(\omega) = \frac{1}{N} \sum\limits_{i=1}^{N} ( Q_{\omega}(s_{i},\widetilde{a_{i}}) - y_{i}) \nabla_{\omega} Q_{\omega} (s_{i},\widetilde{a_{i}})$
		
		\If{t mod d} % If 语句，需要和EndIf对应
		\State Use ($s_{i}$, $a_{i}$, $r_{i}$, $s_{i+1}$) to update actor network $\mu_{\theta}$: 
		\State  
		$\nabla_{\theta} \mathcal J (\theta) = \frac{1}{N} \sum\limits_{i=1}^{N} (\gamma_{D} \nabla_{\widetilde{a}} D_{\phi} (s_{i},\widetilde{a_{i}}) \nabla_{\theta} \mu_{\theta} (s_{i}) + $
		\State \hspace*{1.3in} $ \nabla_{\widetilde{a}} Q_{\omega} (s_{i},\widetilde{a_{i}}) \nabla_{\theta} \mu_{\theta}(s_{i}) )$
		\State $\theta^{\prime} \leftarrow \tau\theta+(1-\tau)\theta^{\prime}$
		\EndIf
		\State \textbf{end if}
		
		\State When t is odd, update target networks:
		\State \hspace*{0.1in} $\omega_{1 \mhyphen 1}^{\prime} \leftarrow \tau \omega_{1}+(1-\tau)\omega_{1 \mhyphen 1}^{\prime}$
		\State \hspace*{0.1in} $\omega_{1 \mhyphen 2}^{\prime} \leftarrow \tau \omega_{1}+(1-\tau)\omega_{1 \mhyphen 2}^{\prime}$
		
		\State When t is even, update target networks:
		\State \hspace*{0.1in} $\omega_{2 \mhyphen 1}^{\prime} \leftarrow \tau \omega_{2}+(1-\tau)\omega_{2 \mhyphen 1}^{\prime}$
		\State \hspace*{0.1in} $\omega_{2 \mhyphen 2}^{\prime} \leftarrow \tau \omega_{2}+(1-\tau)\omega_{2 \mhyphen 2}^{\prime}$
			
		\EndFor
		\State \textbf{end for}
	\end{algorithmic}
\end{algorithm}
%%%%%%%%%%%%%%%%%%%%%%%%%%%%%%%%%%%%%%%%%%%%%%%%%%%%%%%%%%%%%%%%%%%%%%%%

%-------------------------------------------------------------------------
\subsection{Policy network delay update and target policy smoothing regularization}
Bias is introduced in the process of function approximation using neural networks, and the problem of error accumulation will come to the fore as the number of iterations increases. In the existing actor-critic framework, if the critic' assessments are inaccurate, the actor will not learn an accurate policy. In order to make the critic's evaluation more accurate, the frequency of critic updates can be increased and the frequency of actor updates can be decreased, which is the method of delaying the update of the policy network used in the TD3 \cite{fujimoto2018addressing} algorithm. In our proposed actor-director-critic framework, we also adopt this approach, and the director network and the critic network converge earlier than the actor network, so that the actor network can be instructed to reduce unnecessary and inefficient update operations when updating. Ultimately, training efficiency can be improved.

In addition, the TD3 \cite{fujimoto2018addressing} algorithm also uses the processing of adding clipped noise to the target action to solve the problem of error accumulation, with the aim of smoothing the values of a small area around the target action enough so that those actions that are similar to the target action can have similar values to the target action. This smoothing operation based on noise processing can reduce the impact of certain wrong value estimates on the whole policy learning and reduce the generation of errors, so we retain this noise processing operation when improving the TD3 algorithm. This can be described as follows:
\mathchardef\mhyphen="2D
\begin{eqnarray}
	\epsilon \leftarrow clip (\mathcal N (0,\sigma),\mhyphen c, c)
\end{eqnarray}
\begin{eqnarray}
	a=\pi(s)+\epsilon
\end{eqnarray}

%------------------------------------------------------------------------
\section{Proposed method}
\label{sec:formatting}
\subsection{Actor-Director-Critic framework}
There are three roles in actor-director-critic framework: actor, director, and critic. As in actor-critic framework, the critic in the actor-director-critic framework is still a network representing a value function that learns a value function based on the data collected from the actor's interaction with the environment. Critic plays a role in evaluating the long-term benefits of the actor's actions by comparing the magnitude of the value function. Critic's optimization goal can also be expressed as follows:
\begin{eqnarray}
	min  \mathcal L = \mathbb{E} [(Q(s,a) - (r + \gamma Q(s^{\prime},a^{\prime})))^{2}]
\end{eqnarray}

Director is a network that represents the discriminative function. In specific experiments, we classify the actions based on the feedback rewards as simple superiority or inferiority. We obtain a low quality dataset and a high quality dataset, and then train the discriminative ability of the director network on the actions. We introduced a decay factor to make the director work mainly in the initial phase of training. In the initial stage of training, the actor can quickly recognize which actions are worth exploring and which actions should be avoided with the help of the director, and then quickly learn to reduce the exploration of those poorer actions and improve the training speed. The optimization objective of the director is as follows:
\begin{eqnarray}
	\begin{split}
		max  \mathcal V = \mathbb{E}_{a_{h} \sim p_{data}(a_{h})} [D(s,a_{h})] + \\
		\mathbb{E}_{a_{l} \sim p_{data}(a_{l})} [1-D(s,a_{l})]
	\end{split}
\end{eqnarray}

The actor is still a network that represents the policy function and chooses the actions to be performed based on the state information of the current environment. At this point, it no longer receives guidance only from the critic, but also from the director. The critic is the lifelong teacher who teaches the actor to make action decisions in the long-term interest, while the director, by design, is the stage teacher who primarily guides the actor in the early stages of learning to quickly distinguish between the low quality actions and high quality actions. Under the joint guidance of both, the actor learns the optimal policy much faster. $\gamma$ is a decay factor, and actor's optimization goals can be described as follows:
\begin{eqnarray}
	max  \mathcal J= \gamma \mathbb{E} [D(s,a)] + \mathbb{E} [Q(s,a)]
\end{eqnarray}

%-------------------------------------------------------------------------
\subsection{Improved double estimator method}
In the actor-director-critic framework, we design two critic networks and two target critic networks for each critic network. Target critic networks are independent of each other and do not need to be trained. We only need to copy the parameters of the critic network to the target critic network in a soft update manner periodically. It should be noted that for each critic network, the two target critic networks are updated alternately, and only one of the target networks is updated at a time. When optimizing the two critic networks used, they share the same target value. The target value is calculated by first averaging the outputs of the two target critic networks for each critic network and then taking the minimum of the two averages, as shown in the following equation:
\mathchardef\mhyphen="2D
\begin{eqnarray}
\begin{split}
	Q^{\prime}=min (\frac{Q_{1 \mhyphen 1}(s^{\prime},a^{\prime}) + Q_{1 \mhyphen 2}(s^{\prime},a^{\prime})}{2}, \\
	\frac{Q_{2 \mhyphen 1}(s^{\prime},a^{\prime}) + Q_{2 \mhyphen 2}(s^{\prime},a^{\prime})}{2})
\end{split}
\end{eqnarray}

The advantage of this design is that precisely because the alternate updates of the target critic networks make them represent the training results of the critic network in different historical periods, the average of the target values calculated by the training results in different historical periods can better dilute some prediction biases and thus improve the stability and accuracy of the target values.

%-------------------------------------------------------------------------
\subsection{Classified twin delayed deep deterministic policy gradient algorithm}
As mentioned in the introduction, in order to test the performance of the above actor-director-critic framework and the improved double estimator method, we apply them to the TD3 algorithm to improve the TD3 algorithm. The TD3 algorithm is chosen because it is a well-performing algorithm based on the actor-critic framework. The full name of the TD3 algorithm is twin delayed deep deterministic policy gradient algorithm, and we refer to the improved algorithm as classified twin delayed deep deterministic policy gradient algorithm, abbreviated as CTD3. The implementation process of CTD3 is shown in Figure 1.

The actor makes action choices based on the state of the environment it observes. There are two critic networks, and each critic network has two target critic networks. The parameters of the target critic networks are copied from the critic network in the way of soft updating alternately. To solve the overestimation problem, the minimum value of the output of the two critic networks is chosen as a reference to evaluate the actor's action in the long term interest. In addition, we select appropriate reward value cut-off to classify actions into low quality actions and high quality actions based on the reward values returned from the environment, and use low-quality and high-quality datasets to train the discriminative ability of the director. Then director can guide the actor to reduce repetitive exploration of low quality actions during training.

In the CTD3 algorithm we keep the two methods used in the TD3 algorithm to solve the error accumulation problem. One is to delay the update of the policy network, that is, to reduce the frequency of actor updates, and the other is the smoothing operation of the target policy, that is, to add clipped noise. The way to add clipped noise can be described as follows:
\begin{eqnarray}
	\epsilon \leftarrow \mathcal N(0,\sigma)
\end{eqnarray}
\begin{eqnarray}
	\widetilde{a} = \mu_\theta(s)+\epsilon
\end{eqnarray}
\begin{eqnarray}
	\epsilon^{\prime} \leftarrow clip(\mathcal N(0,\sigma^{\prime}),\mbox{-} c,c)
\end{eqnarray}
\begin{eqnarray}
	\widetilde{a^{\prime}} = \mu_{\theta^{\prime}}(s^{\prime})+\epsilon^{\prime}
\end{eqnarray}

The target action of adding truncated noise will be used for the calculation of the target value of the value function. Two target critic networks are designed for each critic network to solve the overestimation problem. 
\mathchardef\mhyphen="2D
\begin{eqnarray}
	\begin{split}
		Q^{\prime}=min (\frac{Q_{\omega_{1 \mhyphen 1}^{\prime}}(s^{\prime},\widetilde{a^{\prime}}) + Q_{\omega_{1 \mhyphen 2}^{\prime}}(s^{\prime},\widetilde{a^{\prime}})}{2}, \\
		\frac{Q_{\omega_{2 \mhyphen 1}^{\prime}}(s^{\prime},\widetilde{a^{\prime}}) + Q_{\omega_{2 \mhyphen 2}^{\prime}}(s^{\prime},\widetilde{a^{\prime}})}{2})
	\end{split}
\end{eqnarray}
\begin{eqnarray}
	y=r+\gamma_Q Q^{\prime}
\end{eqnarray}

The optimization objective of the critic network is to minimize the loss between the predicted target value y and the output value $Q_{\omega}(s,\widetilde{a})$.
\begin{eqnarray}
	\mathcal L (\omega) = \mathbb{E} [(Q_{\omega}(s,\widetilde{a}) - y)^{2}]
\end{eqnarray}
\begin{eqnarray}
	\nabla_{\omega} \mathcal L(\omega) = \frac{1}{N} \sum\limits_{i=1}^{N} ( Q_{\omega}(s_{i},\widetilde{a_{i}}) - y_{i}) \nabla_{\omega} Q_{\omega} (s_{i},\widetilde{a_{i}})
\end{eqnarray}
\mathchardef\mhyphen="2D
\begin{eqnarray}
\begin{split}
	y_{i} = r_{i} + \gamma_{Q} min (\frac{Q_{\omega_{1 \mhyphen 1}^{^{\prime}}}(s_{i}^{\prime},\widetilde{a_{i}^{\prime}}) + Q_{\omega_{1 \mhyphen 2}^{\prime}}(s_{i}^{\prime},\widetilde{a_{i}^{\prime}})}{2}, \\ 
	\frac{Q_{\omega_{2 \mhyphen 1}^{\prime}}(s_{i}^{\prime},\widetilde{a_{i}^{\prime}}) + Q_{\omega_{2 \mhyphen 2}^{\prime}}(s_{i}^{\prime},\widetilde{a_{i}^{\prime}})}{2})
\end{split}
\end{eqnarray}

The director network learns to discriminate between high and low quality actions through different datasets, with the output being 1 when the input is a high quality action and 0 when the input is a low quality action. Director network is designed with reference to the concept of GAN \cite{goodfellow2014generative}, and its goal is to perform the learning of the discriminant function $\mathcal V(\phi)$.
\begin{eqnarray}
	\begin{split}
		\mathcal V(\phi) = \mathbb{E}_{a_{h} \sim p_{data}(a_{h})} [D_\phi (s,a_{h})] + \\ \mathbb{E}_{a_{l} \sim p_{data}(a_{l})} [1-D_\phi (s,a_{l})]
	\end{split}
\end{eqnarray}
\begin{eqnarray}
	\begin{split}
		\nabla_{\phi} \mathcal V(\phi) = \frac{1}{N} \sum\limits_{i=1}^{N} (\nabla_{\phi} D_{\phi} (s_{i},a_{hi}) - \nabla_{\phi} D_{\phi} (s_{i},a_{li}))
	\end{split}
\end{eqnarray}

The actor network performs policy optimization with the joint help of the critic network and the director network, and the optimization goal is to maximize the value of the value function output by the critic network as well as to maximize the value of the discriminant function output by the director network. Here the decay factor $\gamma_{D}$ is used to make the director focus on the early stage of training and speed up the training.

\begin{eqnarray}
	\mathcal J (\theta) = \gamma_{D} \mathbb{E} [D_{\phi}(s,\widetilde{a})] + \mathbb{E} [Q_{\omega}(s,\widetilde{a})]
\end{eqnarray}
\begin{eqnarray}	
	\begin{split}
		\nabla_{\theta} \mathcal J (\theta) = \frac{1}{N} \sum\limits_{i=1}^{N} (\gamma_{D} \nabla_{\widetilde{a}} D_{\phi} (s_{i},\widetilde{a_{i}}) \nabla_{\theta} \mu_{\theta}(s_{i}) + \\
		\nabla_{\widetilde{a}} Q_{\omega} (s_{i},\widetilde{a_{i}}) \nabla_{\theta} \mu_{\theta}(s_{i}) )
	\end{split}
\end{eqnarray}

CTD3 is summarized in Algorithm 1.

%------------------------------------------------------------------------
\section{Experiment}

\subsection{Experimental environments}
\begin{figure}[hbtp]
	\centering
	\includegraphics[width=0.89\linewidth]{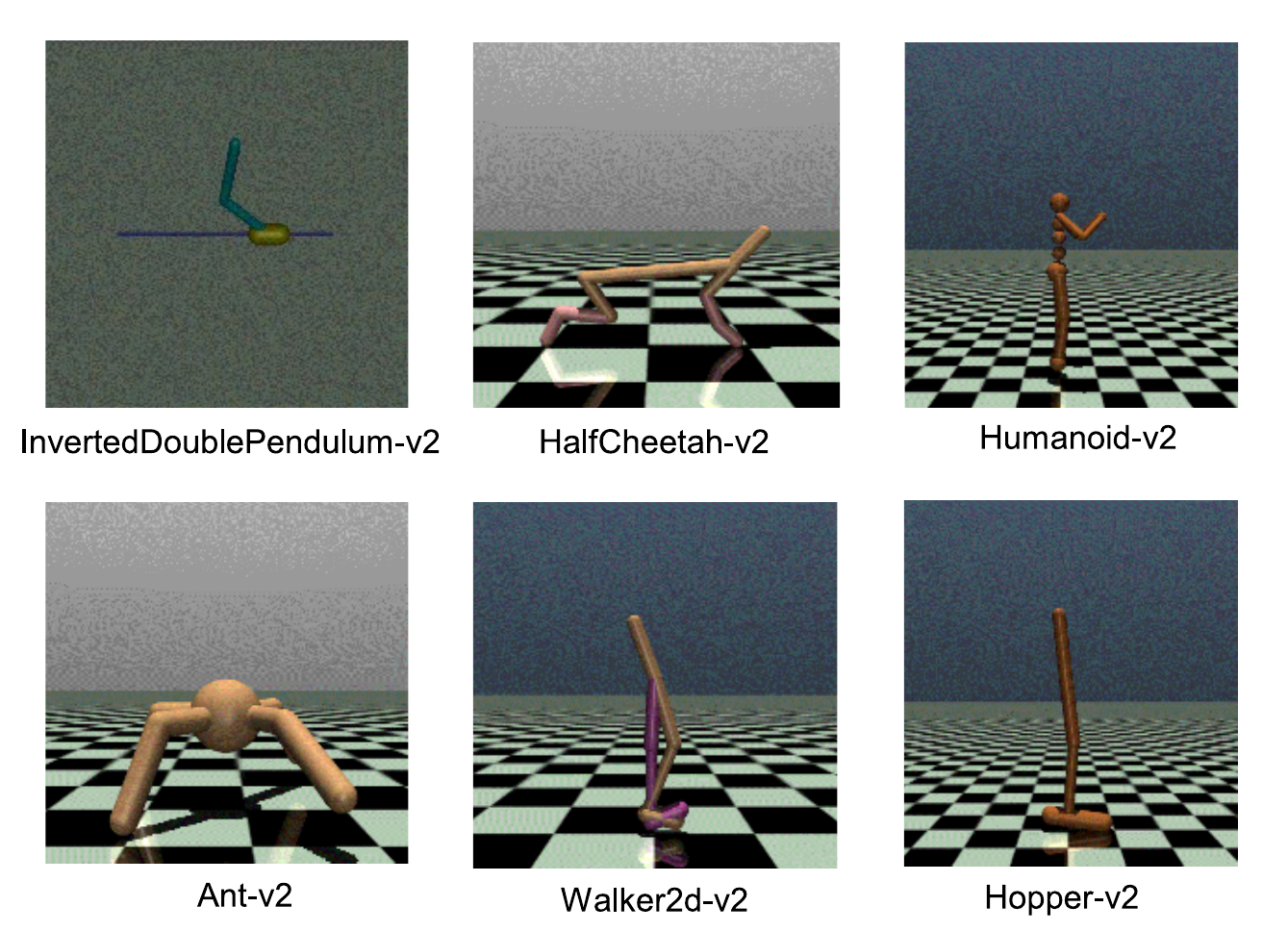}
	\caption{Six experimental environments in MuJoCo.}
	\label{fig:short}
\end{figure}

\begin{figure*} [t]
	\centering
	\includegraphics[width=0.3\linewidth]{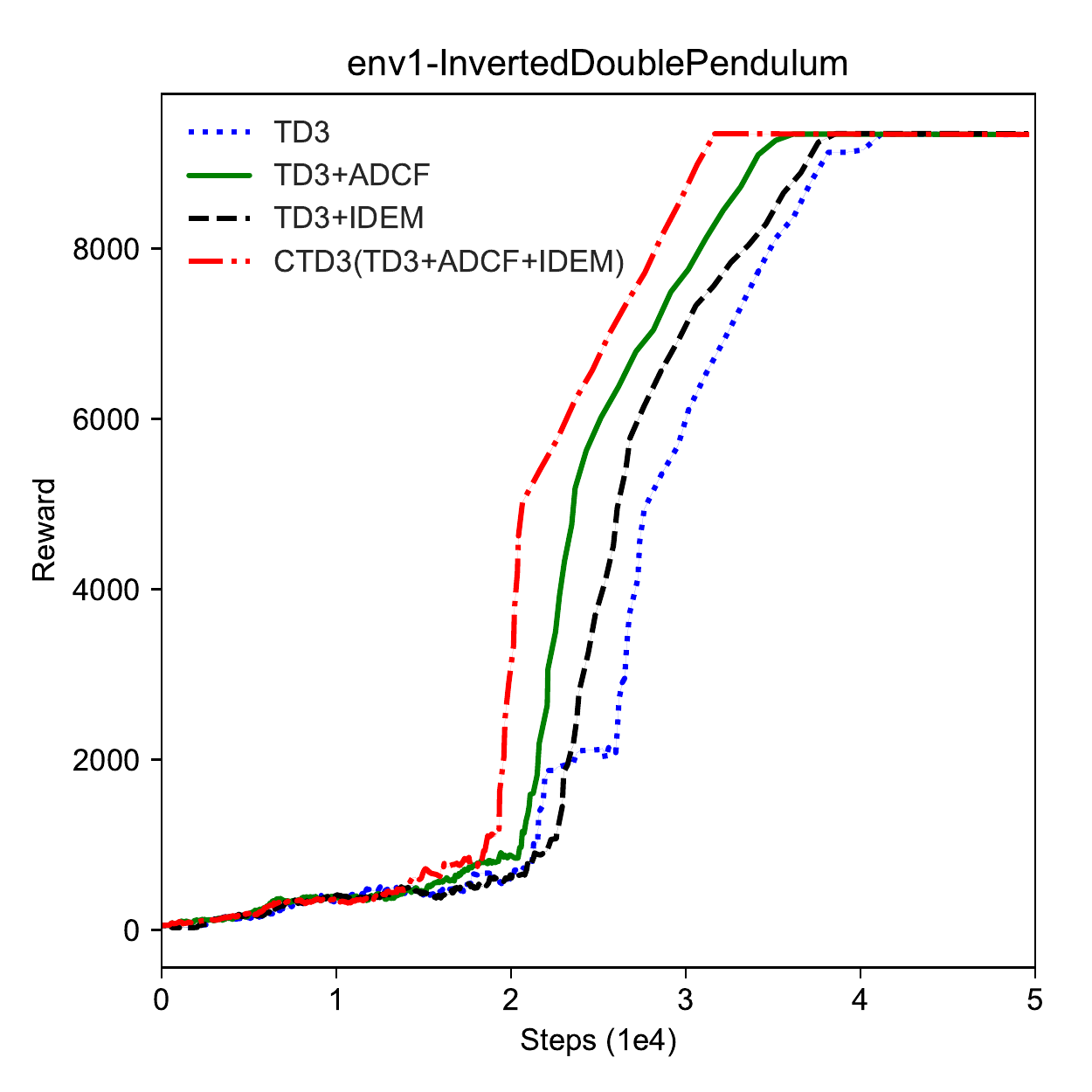} \hspace{-3mm}
	\centering
	\includegraphics[width=0.3\linewidth]{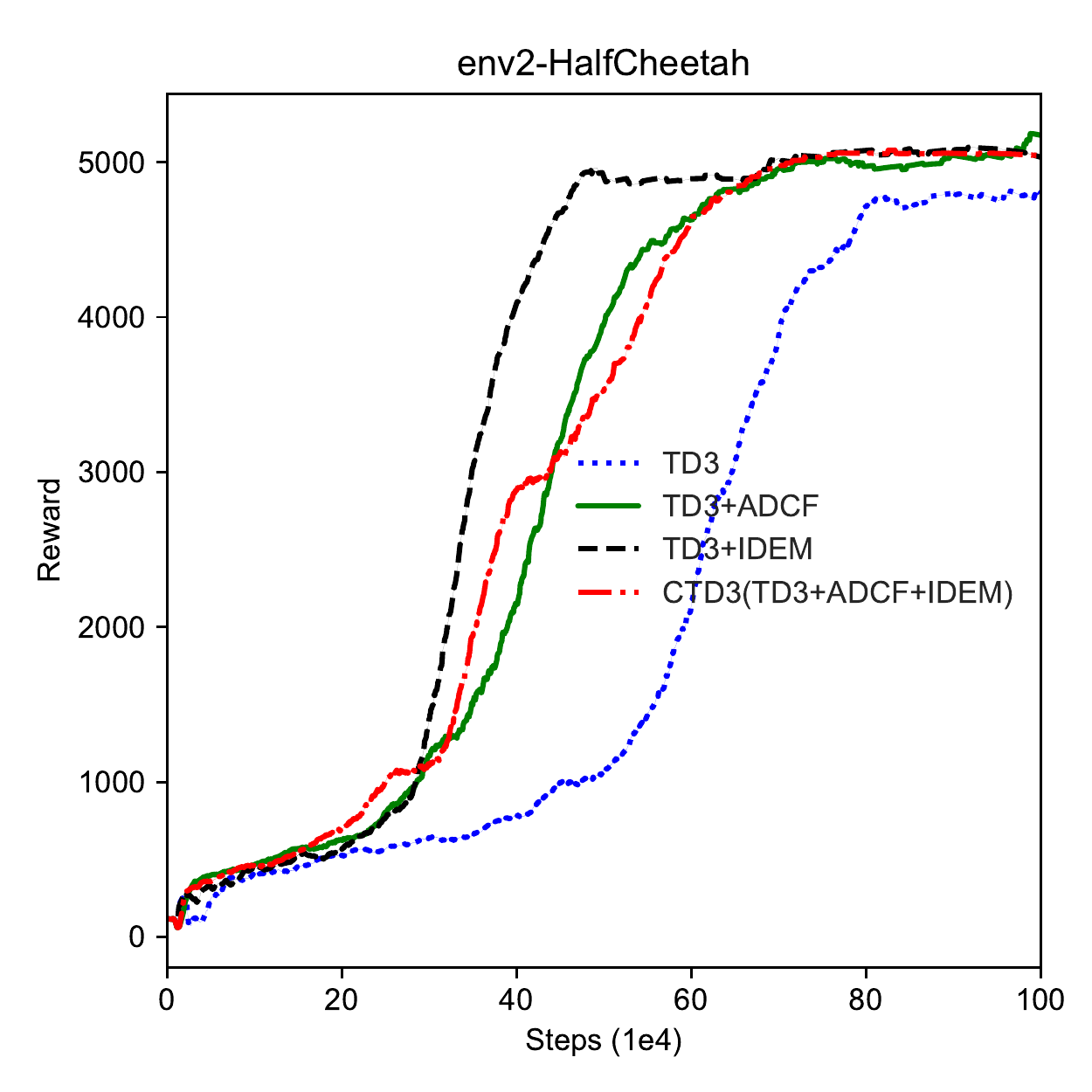} \hspace{-3mm}
	\centering
	\includegraphics[width=0.3\linewidth]{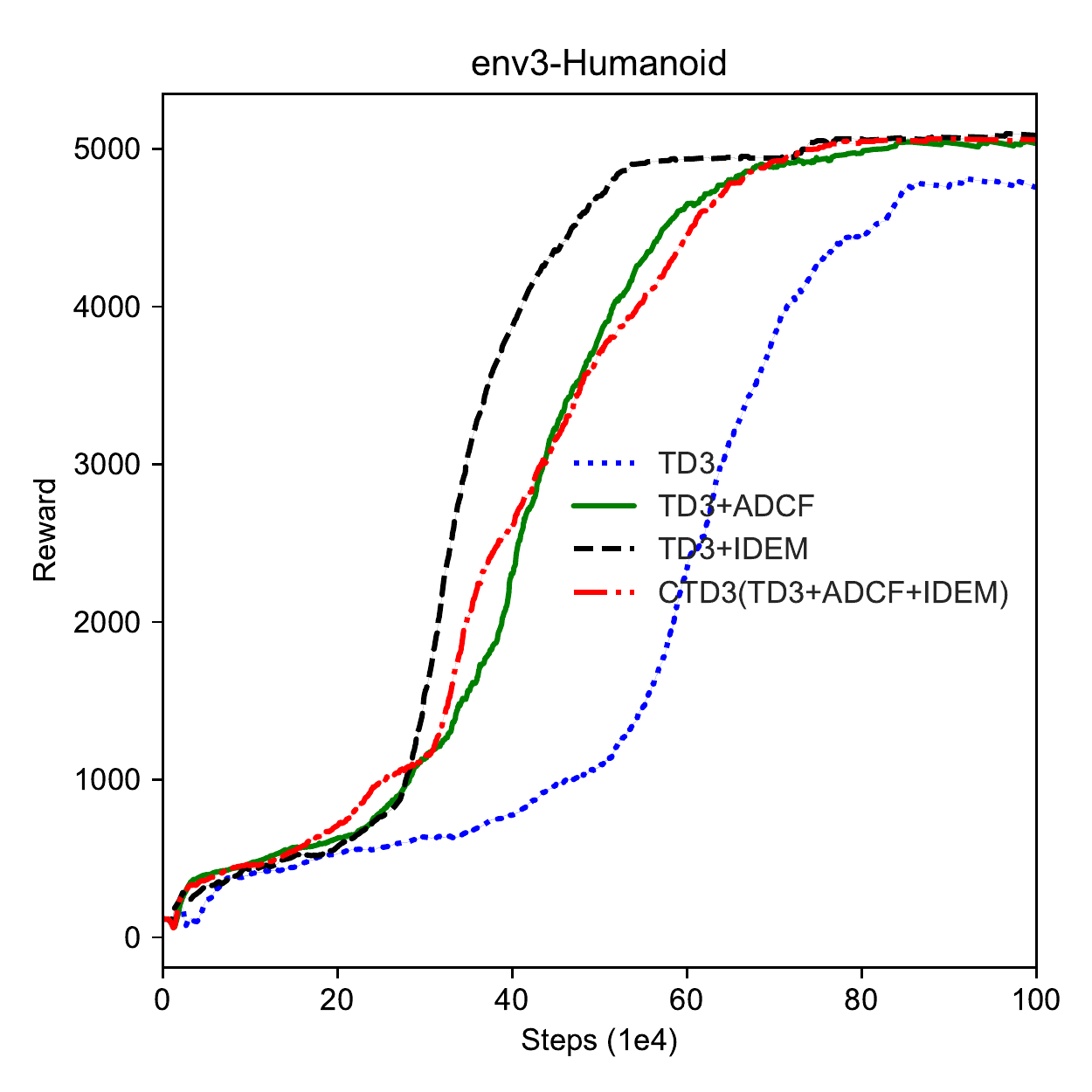} 
	\\
	\centering
	\includegraphics[width=0.3\linewidth]{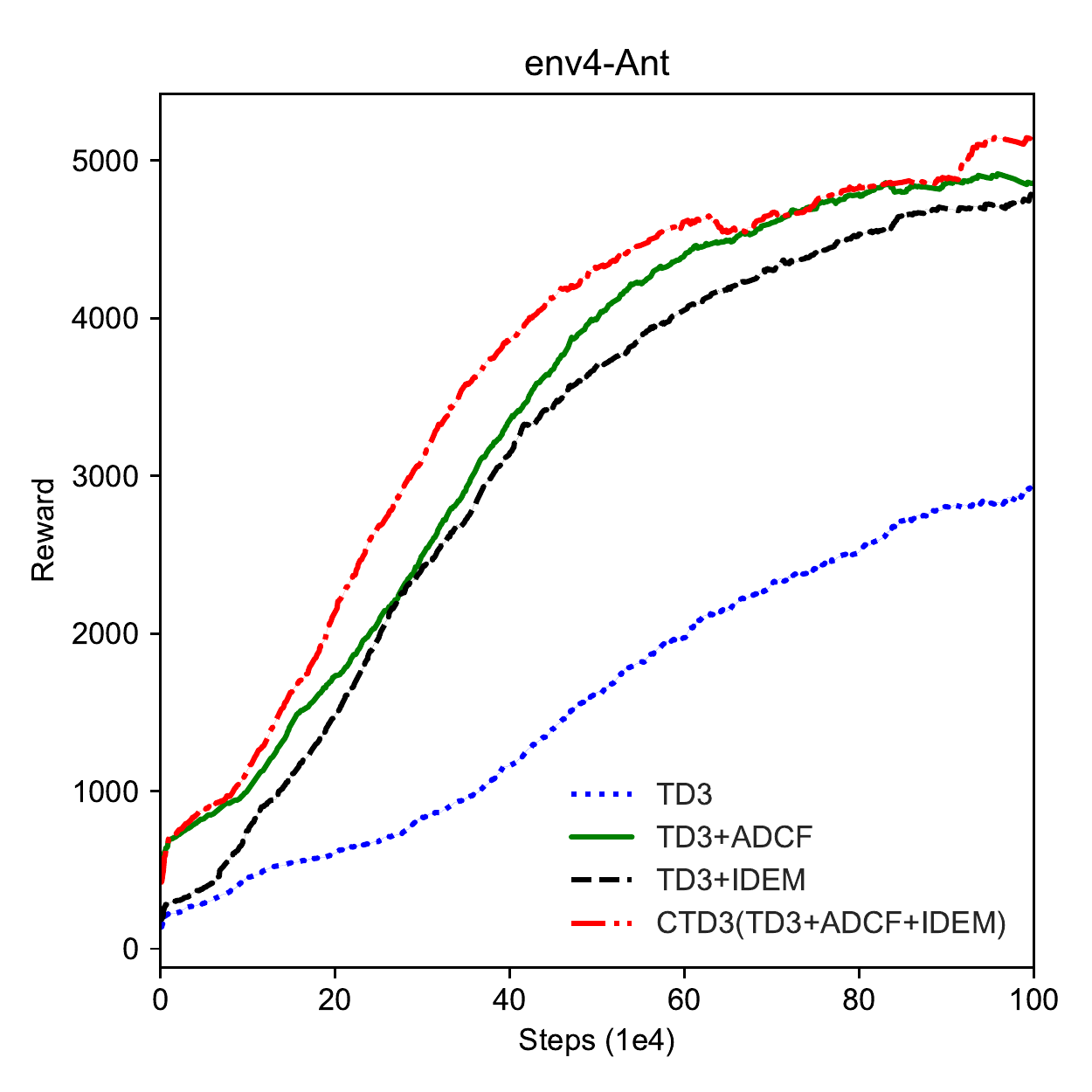} \hspace{-3mm}
	\centering 
	\includegraphics[width=0.3\linewidth]{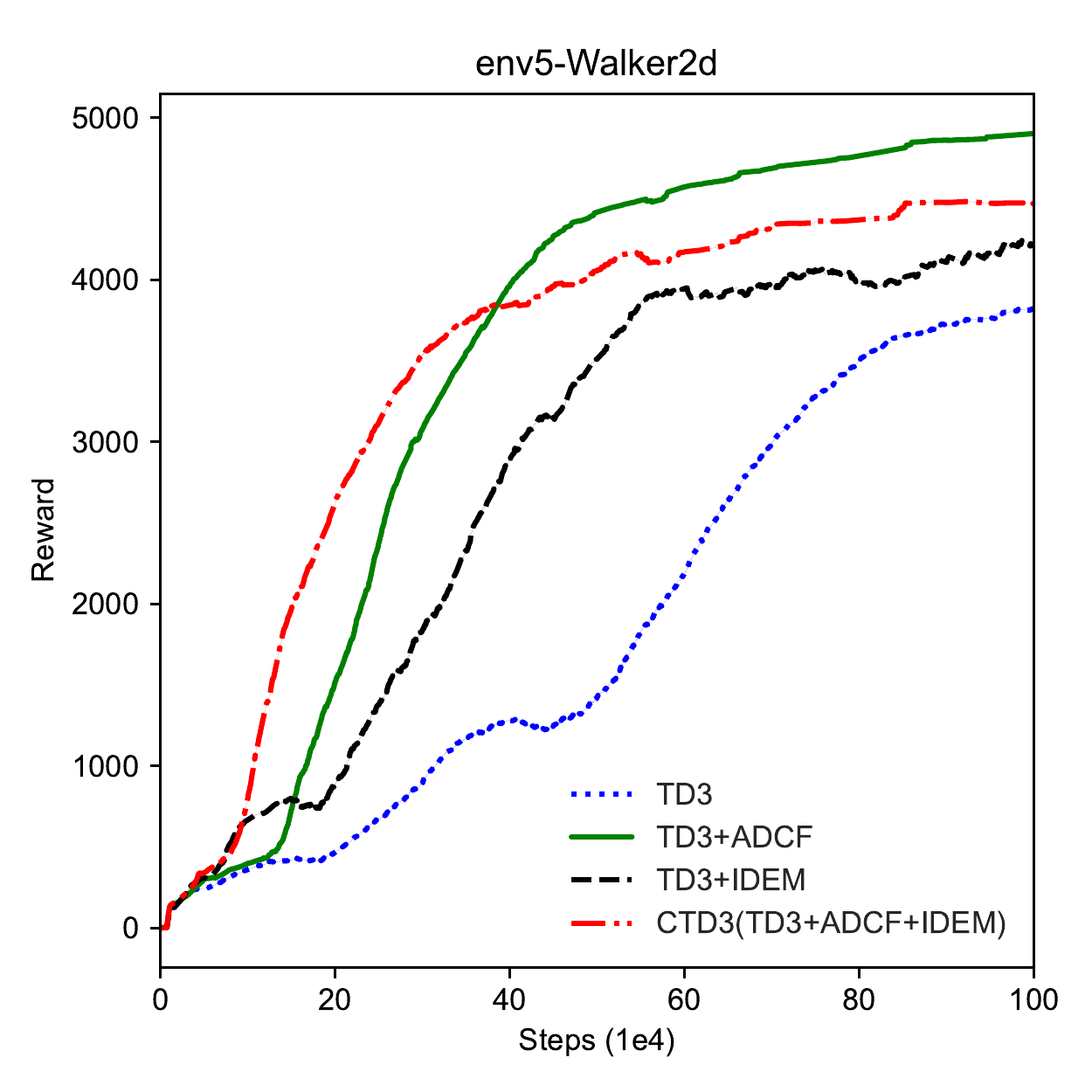} \hspace{-3mm}
	\centering 
	\includegraphics[width=0.3\linewidth]{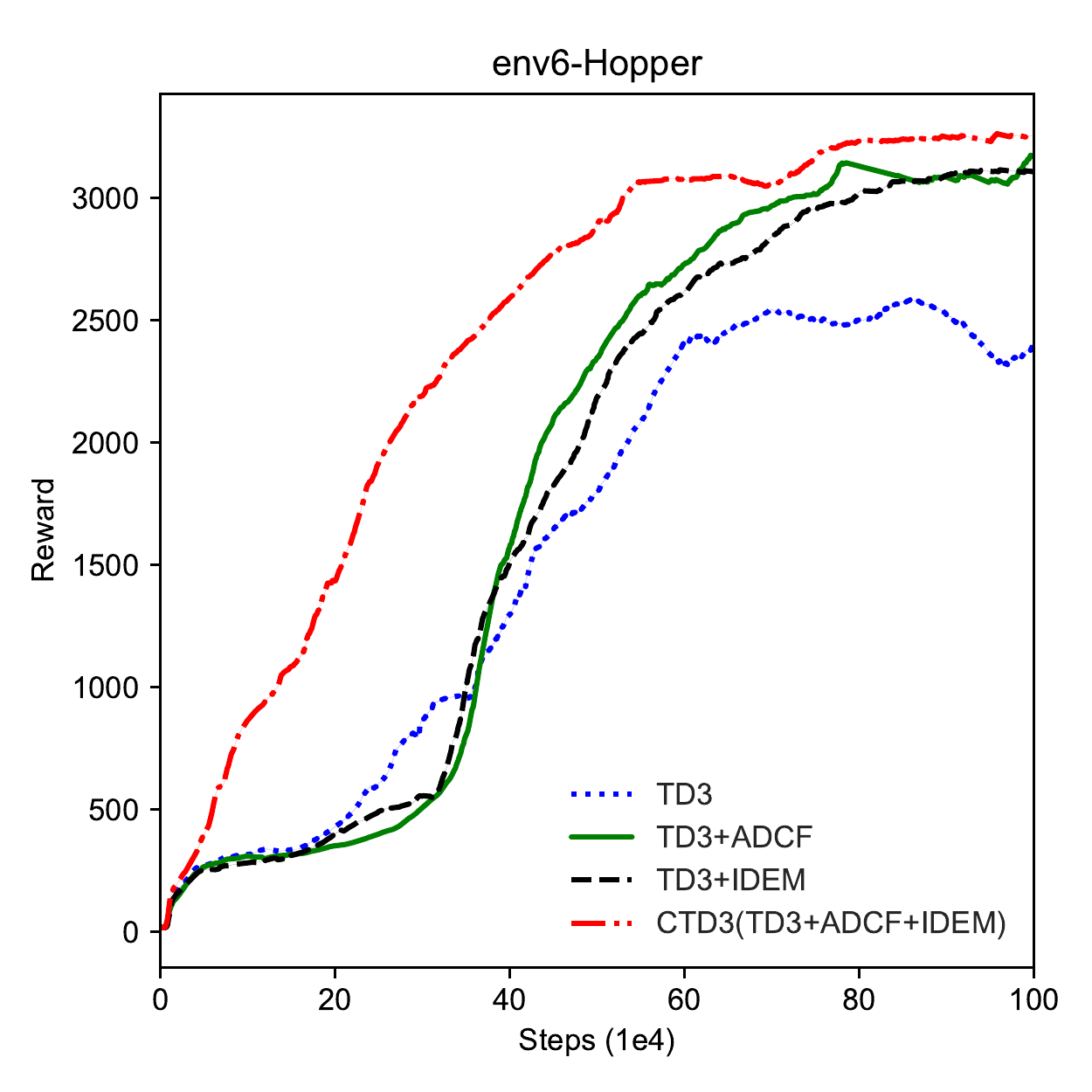}
	\caption{Performance curves in six environments. }
	\label{fig:short} 
\end{figure*}

\begin{table*}[t]
	\centering
	\begin{tabular}{@{}lcccc@{}}
		\toprule
		Environment & TD3 & TD3+ADCF & TD3+IDEM & CTD3(TD3+ADCF+IDEM)  \\
		\midrule
		InvertedDoublePendulum & 9341.302 & 9342.752 & 9350.902 & 9342.690  \\
		HalfCheetah & 8214.770 & 9460.794 & 9156.181 & 10324.831  \\
		Humanoid & 4755.148 & 5034.893 & 5085.446 & 5057.048  \\
		Ant & 2949.150 & 4861.062 & 4779.486 & 5171.217  \\
		Walker2d & 3818.958 & 4902.341 & 4208.518 & 4468.348  \\
		Hopper & 2395.742 & 3169.502 & 3118.340 & 3252.941  \\
		\bottomrule
	\end{tabular}
	\caption{Experimental results in six environments.}
	\label{tab:example}
\end{table*}

We conducted experiments in MuJoCo simulation environment. The six experimental environments selected in Figure 2 are all continuous action environments and are rich in reward value settings.

\subsection{Comparative evaluation and ablation study}
As already mentioned, in order to test the effectiveness of the proposed actor-director-critic framework and the improved double estimator method, we apply them to the TD3 algorithm. CTD3 algorithm is obtained by improving TD3. First of all, we would like to explain why we chose TD3 algorithm for comparison, because TD3 is a very good reinforcement learning algorithm based on actor-critic framework. In the original paper of the TD3 \cite{fujimoto2018addressing} algorithm published in ICML, this method can completely beat DDPG, PPO, TRPO, ACKTR, and SAC. Therefore, we choose the TD3 algorithm for comparison experiments.

We conduct comparative experiments on CTD3 and TD3 in six environments under the MuJoCo platform. For each environment a specific random seed is set and the maximum number of experimental steps is limited. The data are smoothed at the end of the experiments, using the sliding average filtering method. A suitable sliding window size is set for each data of different environments, and the data in the window are averaged. All data are smoothed when the window is sliding from the beginning to the end.
On the one hand, the smoothed data are written to Table 1, and it should be noted that the data of the six environments written to Table 1 are the data at the end of the last training round run by agent, both for CTD3 and TD3. On the other hand, all the smoothed data of CTD3 and TD3 are plotted as curves to obtain Figure 3, where the training process of CTD3 and TD3 in each environment can be visually compared. 

CTD3 is a new algorithm that improves TD3 by applying the two ideas of actor-director-critic framework and the improved double estimator method proposed in this paper to the TD3 algorithm. We can compare the performance of CTD3 and TD3 by comparing the data in Table 1 and the curves in Figure 3. In all the six environments, the total return of CTD3 is higher than that of TD3, and the convergence rate of CTD3 is also faster than that of TD3.
Since the performance of the CTD3 algorithm is better than TD3 in terms of convergence speed and final score, it can prove that the actor-director-critic framework and the improved double estimator method proposed in this paper are feasible.
%%%%%%%%%%%%%%%%%%%%%%%%%%%%%%%%%%%%%%%%%%%%%%%%%%%%%%%%%%%%

TD3+ADCF means that the ADC framework is applied to the TD3 algorithm to improve the TD3 algorithm.
TD3+IDEM means that the improved double estimator method is applied to the TD3 algorithm to improve the TD3 algorithm.
CTD3(TD3+ADCF+IDEM) means that the ADC framework and the improved double estimator method are applied to the TD3 algorithm at the same time to improve the TD3 algorithm and obtain the CTD3 algorithm.
For ablation experiments, we can analyze the performance of the ADC framework and the improved double estimator method proposed in this paper by comparing the data of TD3, TD3+ADCF, TD3+IDEM and CTD3 in Table 1 and the curves in Figure 3.
As shown in Table 1, in all six experimental environments, no matter TD3+ADCF or TD3+IDEM, their total returns are higher than those of TD3. The curve in Figure 3 also shows that in all six experimental environments, no matter TD3+ADCF or TD3+IDEM, their convergence speed is faster than that of TD3.
In most cases, CTD3 outperforms TD3+ADCF and TD3+IDEM, but sometimes it is outperformed by one or the other.

In conclusion, the improved performance compared to TD3, either using the actor-director-critic framework alone or the improved double estimator method alone, or use both of them, are sufficient to demonstrate the feasibility of our proposed methods in this paper.

\section{Conclusion}

Reinforcement learning is a kind of trial-and-error learning, which has the problem of low sample utilization. Moreover, in existing deep reinforcement learning algorithms based on the actor-critic framework, there is also a problem that it is difficult for an agent to rely on the policy function to make good actions until the value function is learned to the extent that it can provide reasonable guidance to the policy function. Therefore, in order to alleviate these two problems, this paper proposes a new deep reinforcement learning framework, actor-director-critic framework. The director role is added to the actor-critic framework, and a decay factor is set for the director. By training the director to discriminate between low quality and high quality actions, it can help the actor reduce repetitive attempts to perform low quality actions during exploration, thus improving sample utilization and speeding up training.

In solving the problem of overestimation in reinforcement learning, for the method called double estimator method that requires training two Q networks and designing one target network for each Q network, this paper tries to propose an improved double estimator method to design two target networks for each Q network. 
For each Q network, its two target networks are alternately soft updated. The outputs of these target networks are comprehensively used to calculate the target value of Q network. First, the output average of the two target networks owned by each Q network is calculated, and then the smaller of the two average values is taken as the final target value of the Q network, and this target value is shared by both Q networks. so as to alleviate the problem of overestimation. So that the problem of overvaluation can be mitigated much better than before.

To test the feasibility of both the actor-director-critic framework and the improved double estimator method, we applied them to the TD3 algorithm and improved the TD3 algorithm to obtain the CTD3 algorithm. We conducted comparative and ablation experiments in several environments in the MuJoCo. Experimental results show that the CTD3 algorithm generally has faster convergence speed and higher total return than TD3, proving the feasibility of the methods proposed in this paper.

%%%%%%%%% REFERENCES
\balance
{\small
\bibliographystyle{ieee_fullname}
\bibliography{egbib}
}

\end{document}